\pdfoutput=1

\documentclass[11pt]{article}

\usepackage{EMNLP2023}

\usepackage{times}
\usepackage{latexsym}
\usepackage{graphicx}
\usepackage{algorithm}
\usepackage{algpseudocode}
\usepackage{amsmath}
\usepackage{colortbl}
\usepackage{makecell}
\usepackage{booktabs}
\usepackage{subcaption}
\usepackage{ulem}

\usepackage[T1]{fontenc}

\usepackage[utf8]{inputenc}

\usepackage{microtype}

\usepackage{inconsolata}

\title{Handling Realistic Label Noise in BERT Text Classification}

\author{Maha Tufail Agro \\
  MBZUAI \\
  \texttt{maha.agro@mbzuai.ac.ae} \\\And
  Hanan Aldarmaki \\
  MBZUAI \\
  \texttt{hanan.aldarmaki@mbzuai.ac.ae} \\}

\begin{document}
\maketitle
\begin{abstract}
Label noise refers to errors in training labels caused by cheap data annotation methods, such as web scraping or crowd-sourcing, which can be detrimental to the performance of supervised classifiers. Several methods have been proposed to counteract the effect of random label noise in supervised classification, and some studies have shown that BERT is already robust against high rates of randomly injected label noise. However, real label noise is not random; rather, it is often correlated with input features or other annotator-specific factors. In this paper, we evaluate BERT in the presence of two types of realistic label noise: feature-dependent label noise, and synthetic label noise from annotator disagreements. We show that the presence of these types of noise significantly degrades BERT classification performance. To improve robustness, we evaluate different types of ensembles and noise-cleaning methods and compare their effectiveness against label noise across different datasets. 

\end{abstract}

\section{Introduction}
Deep learning algorithms have been shown to perform extremely well in supervised classification tasks given high-quality datasets. Unfortunately, obtaining gold-standard labels is often prohibitively expensive with large-scale datasets, leading practitioners to resort to cheaper data collection methods such as crowd-sourcing or automatic annotation methods \cite{annotators2014}. These techniques are known to impart a substantial amount of label noise in the data, which can degrade classification performance \cite{noise_degradation}. Label noise refers to errors or inconsistencies within the data labels, such that the prescribed labels do not match the gold labels assigned by experts. Datasets obtained through web scraping often contain label noise given the absence of expert-verified gold labels \cite{distilation}. 
Due to a meteoric rise in social media usage, more and more datasets are automatically acquired from online social platforms, and such datasets are likely to contain label noise. Small-scale datasets can also suffer from the same problem if the annotation process is challenging or the annotators have divergent opinions \cite{Ma2019BlindIQ}. 

Some prior works have been dedicated to developing and deploying algorithms that combat the effects of label noise in text classification \cite{han2018coteaching,2014sukhbataar,zhang2018,jiang2018mentornet}.
However, most previous studies simulated label noise by random substitution, and recent research has shown empirically that many methods that successfully handle random noise are ineffective against real-world label noise \cite{synthetic_noise}.
In the text classification domain, \citet{bert2022robust} explored the robustness of previously proposed methods for handling label noise, including noise matrix with regularization \cite{jindal-etal-2019-effective}, co-teaching \cite{han2018coteaching}, and label smoothing \cite{szegedy2016}. They concluded that BERT \cite{Devlin2019BERTPO} is already robust against randomly injected label noise and these approaches obtain no additional performance gains. On the other hand, they find that feature-dependent label noise, which realistically arises from automatic annotation techniques, degrades BERT performance and these noise handling techniques add little to no robustness at all. This creates a need for a comprehensive evaluation of noise-robust methods in the domain of text classification, considering the presence of realistic labeling errors. 

In this paper, we describe methods and experiments for handling realistic label noise in BERT text classification. We use two datasets that contain feature-dependent label noise from automatic annotation, namely Yor\`ub\'a and Hausa \cite{hausa_and_yoruba}. These two datasets have been manually cleaned, so a clean version of each exists for evaluation. In addition, we use tweetNLP \cite{DBLP:conf/acl/GimpelSODMEHYFS11} and SNLI \cite{snli:emnlp2015} datasets with synthetic noise that mimics human errors by utilizing multiple crowd-sourced annotations \cite{chong-etal-2022-detecting}. This collection of datasets provides a range of noise types and levels that more closely reflect realistic label noise compared to random noise injection. We evaluate the performance of vanilla BERT compared with a subset of noise-handling approaches, namely co-teaching \cite{han2018coteaching}, Consensus Enhanced Training Approach (CETA) \cite{liu-etal-2022-ceta}, different types of ensembles \cite{ensemble2022105151}, and noise cleaning \cite{chong-etal-2022-detecting,sluban2014ensemble}.
We summarize our findings as follows:
\begin{enumerate}
\item For datasets with feature-dependent label noise, we find that CETA, some types of ensembles, and noise cleaning, all provide positive performance gains compared to vanilla BERT.  
\item For synthetic label noise from multiple annotations, we do not observe significant gains using these approaches. We surmise that this type of noise is more challenging or may even reflect inherently ambiguous labels.
\end{enumerate}

It is worth noting that the noise is qualitatively different in these two categories of label noise as the latter arises from human rather than automatic processes, which could be due to either errors or genuine disagreements. Some recent works attempt to include multiple labels in the training process rather than rely on a single gold label to account for the inherent uncertainty from human disagreements. This may be justified given the nature of some tasks, and the noising scheme performed on tweetNLP and SNLI may warrant that kind of treatment or further scrutiny to identify clear-cut errors. However, as we focus on noise robustness as the scope of this work, we treat the synthetic noise int these datasets as labeling errors and leave any further analysis of this sort for future work. 

\section{Background \& Related Works}  

\subsection{Types of Label Noise}

Label noise refers to irregularities or inconsistencies within the data labels, where the prescribed label of a data point does not correspond to the true expert label. In other words, noisy instances in this context specifically pertain to inaccuracies or errors in the labeling of the data, rather than any distortions or imperfections in the input data itself. 
 
When observing the effect of label noise, the majority of existing literature in text classification assumes random injection of label noise \cite{han2018coteaching,2014sukhbataar,zhang2018}. This type of synthetic noising involves randomly permuting a fixed number of labels according to a pre-defined noise level and noise type.
Because the process of simulating such noise is entirely random and does not depend on the input data features in any way, this type of noise is also known as \textit{feature-independent}  label noise.

In contrast, \textit{feature-dependent} label noise is correlated with input features \cite{algan2020label}.  
Datasets that use distantly or weakly supervised methods to generate labels are prone to this type of label noise. These approaches are often used in low-resource applications where it is impractical or expensive to manually annotate large amounts of data. Relation extraction is one such application that heavily relies on automatic data generation methods as supervised relation extraction methods necessitate an extensive amount of labeled training data \cite{mintz-etal-2009-distant}. In this area, denoising methods such as the ones proposed in \citet{jia-etal-2019-arnor}, \citet{qin-etal-2018-dsgan}, \citet{liu-etal-2022-ceta} and \citet{ma-etal-2021-sent} are specifically developed to address feature-dependent label noise in relation extraction datasets.

Recently, \citet{chong-etal-2022-detecting} developed realistic noising methods that mimic how humans make labeling errors by taking advantage of the multiple rounds of annotation that some datasets undergo. During the annotation process, certain subsets of the data are subjected to rigorous validation schemes, such as gold labels assigned by experts, while others are annotated using less stringent methods, such as crowdsourced evaluations. By incorporating varying annotations generated during this process, their approach produces realistic label noise that reflects how humans make errors. We refer to this noising scheme as \textit{pseudo-real-world label} noise.
\if{FALSE}
\begin{figure}[ht]
  \centering
  \includegraphics[width=0.9\linewidth]{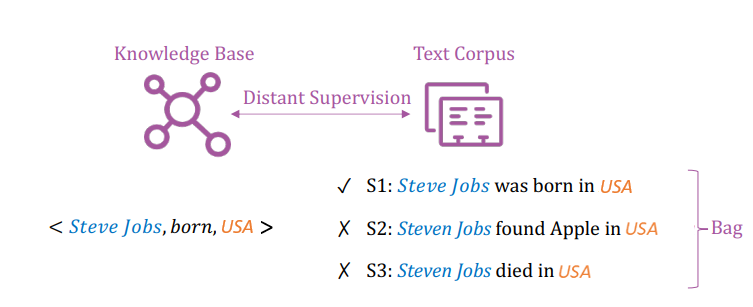}
  \caption{An example of the distant supervision annotation process. Source: \cite{liu-etal-2022-ceta}}
  \label{fig:distant_supervision_noise}
\end{figure}
\fi

\subsection{Noise-robust methods}
Noise-robust methods in the literature include model enhancements such as \textbf{robust loss functions}. Robust loss functions are a class of loss functions used to train models in a way that is more resistant to label noise. One such loss function is the family of generalized cross-entropy loss functions \cite{zhang2018}, which  are designed to be more robust to label noise by penalizing the model less for incorrect predictions that are consistent with noisy labels. 

Another class of noise-robust approaches is what we refer to as \textbf{multi-netowrk training}. This sub-category of methods introduces multiple networks that learn from each other and as such make more informed decisions regarding which data to use to update the model parameters. For instance, co-teaching \cite{han2018coteaching} includes two models that are trained in parallel, and each model is presented with examples that incur low loss by the other model. Intuitively, correct labels produce small losses in earlier training epochs and noisy labels produce higher losses. Similarly, the Consensus-Enhanced Training Approach (CETA) proposed in \citet{liu-etal-2022-ceta} is a methodology for robust sentence-level relation extraction that emphasizes the selection of clean data points during the training process. The denoising technique is applied to establish a robust boundary for classification, preventing inaccurately labeled data from being assigned to the wrong classification space, and the consensus between two divergent classifiers is used to select clean instances for training. 
\subsection{Noise cleaning approaches}

\textit{\textbf{Noise-cleaning}} aims to automatically segregate clean data from noisy data in order to train the final classifier using a cleaned subset of the original training set. The ``small loss trick" is commonly used to identify potentially noisy or mislabeled data. The intuition behind this approach is that noisy data have comparatively higher loss than clean data \cite{Takeda2021,han2018coteaching,jiang2018mentornet,ji2021}. 

Several approaches have been proposed for automatic noise detection, which can be a first step towards noise-cleaning before training a robust classifier. \citet{wheway2001boosting} used boosting to detect noisy data instances. The approach involves iteratively re-weighing the data points to emphasize those that are most difficult to classify correctly. The resulting model is then used to identify the noisy data points by measuring their contribution to the final model. \citet{sluban2014ensemble} trained multiple classifiers (ensemble) on different subsets of the data and combined their outputs to obtain a noise ranking for each instance. Similarly, \citet{chong-etal-2022-detecting} assessed the performance of pre-trained language models as error detectors using clean held-out data. They experiment with the error detection capabilities of individual pre-trained models and an ensemble of pre-trained language models. They find that an ensemble of pre-trained model losses outperforms individual model loss in error detection. 

\subsection{Label noise \& BERT}
BERT \cite{Devlin2019BERTPO} is a popular pre-trained language model that is frequently used for text classification by fine-tuning on target labels. Some recent studies have shown that BERT is already robust against randomly injected label noise \cite{bert2022robust}, and early stopping is sufficient to prevent overfitting on noisy labels. \citet{bert2022robust} evaluates popular noise robust approaches in BERT text classification such as appending noise transition matrix after BERT's predictions \cite{2014sukhbataar}, acquiring the noise transition matrix with $l2$ regularization \cite{jindal-etal-2019-effective}, and multi-network training via co-teaching \cite{han2018coteaching}. They conclude that while BERT appears to be inherently robust to feature-independent noise, none of the above approaches improves BERT's peak performance in the presence of feature-dependent label nose.

\section{Methodology}
In this work, we evaluate BERT text classification on datasets containing pseudo-real-world label noise and feature-dependent label noise. We do not consider randomly injected label noise as \citet{bert2022robust} have shown BERT to be already robust to this type of synthetic noise. The scope of this work is limited to text classification with BERT following the baselines established by \citet{bert2022robust}. 

\subsection{Datasets}

\begin{table*}[ht]
\centering
\begin{tabular}{|l|c|c|c|c|}
\hline
\rowcolor{gray!50}
Dataset & Yor\`ub\'a & Hausa & TweetNLP & SNLI \\
\hline
Number of classes & 7 & 5 & 15 & 3 \\
Average sentence length & 13 & 10 & 12 & 21 \\
Train Samples & 1340 & 2045 & 11565 & 363043 \\
Validation Samples & 189 & 290 & 2874 & 9831 \\
Test Samples & 379 & 582 & - & 9815 \\
Train Noise Level & 33.28\% & 50.37\% & Various & Various \\
\hline
\end{tabular}
\caption{Dataset statistics}
\label{table:data-statistics}
\end{table*}

To study feature-dependent label noise, we use two news-topic categorization datasets from two low-resource African languages: Hausa and Yor\`ub\'a \cite{hausa_and_yoruba}. These languages are spoken by large populations in Africa, with Hausa being the second most spoken indigenous language, with 40 million native speakers, and Yor\`ub\'a being the third most spoken, with 35 million native speakers{\footnote{https://en.wikipedia.org/wiki/Languages\_of\_Africa}}. For these datasets, gazetteers were used for automatic labeling, which results in feature-dependent label noise. For instance, when identifying texts for the ``Africa" class, a labeling rule based on a list of African countries and their capitals was employed. These datasets were chosen specifically as they contain automatic annotation label noise i.e., weak-supervision/feature-dependent noise in addition to clean versions of the splits, making it possible to establish ground truth. Note that the amount of label noise in Hausa and Yor\`ub\'a is fixed. 

Furthermore, we use the noising schemes proposed by \citet{chong-etal-2022-detecting} to simulate real-world label noise produced by crowd-sourced labeling. Pseudo-real-world label noise is injected in two benchmark datasets: TweetNLP \cite{DBLP:conf/acl/GimpelSODMEHYFS11} and Stanford Natural Language Inference (SNLI) dataset \cite{snli:emnlp2015}. TweetNLP is a part-of-speech tagging dataset developed by scraping Twitter posts. While TweetNLP already contained crowd-sourced labels, it later received separate crowdsource evaluations, allowing access to multiple annotations from separate annotators. The SNLI dataset is a large Natural Language Inference corpus developed at Stanford. The original corpus consists of 570K sentence pairs, manually labeled by experts. Like TweetNLP, a subset of SNLI later received extensive crowdsource evaluation. We noise both TweetNLP and SNLI to three label noise levels: 10\%, 20\%, and 30\%. Data statistics for all datasets are shown in Table \ref{table:data-statistics}.

 \if{False}
\begin{table}[!h]
\centering
\scalebox{0.60}{
\begin{tabular}{lcccccc}
\hline
\hline
\rowcolor{gray!50}
Dataset & Classes & \makecell{Average\\Lengths} & \makecell{Train\\Samples} & \makecell{Validation\\Samples} & \makecell{Test\\Samples} & \makecell{Train\\Noise\\Level} \\
\hline
\hline
Yor\`ub\'a & 7 & 13 & 1340 & 189 & 379 & 33.28\% \\
Hausa & 5 & 10 & 2045 & 290 & 582 & 50.37\% \\
TweetNLP & 15 & 12 & 11565 & 2874 & - & Various \\
SNLI & 3 & 21 & 363043 & 9831 & 9815 & Various \\
\hline
\hline
\end{tabular}
}
\caption{Dataset statistics.}
\label{table:data-statistics}
\end{table}
\fi


\subsection{Baselines}
\citet{bert2022robust} already evaluated the noise matrix approach \cite{2014sukhbataar}, label smoothing \cite{szegedy2016}, and co-teaching \cite{han2018coteaching} on the feature-dependent datasets, Hausa and Yor\`ub\'a,  and concluded that no gains are observed using these methods. We use the following as baselines to benchmark our experiments using other approaches: 
\begin{enumerate}
    \item \textbf{Vanilla BERT}: BERT trained on noisy training data without a noise-handling mechanism, except early stopping on a noisy validation set, as done in \citet{bert2022robust}.
    \item \textbf{Co-teaching} \cite{han2018coteaching}, which simultaneously trains two networks, with each network independently ranking data points based on their loss to guide the other network on which points to be included for training. In other words, each network independently performs noise-cleaning for the other network.
   
\end{enumerate} 

\section{Approaches}

We experiment with the following approaches as potential methods for improving performance under realistic label noise conditions:
\subsection{Consensus-Enhanced Training Approach (CETA)}

CETA \cite{liu-etal-2022-ceta} has been proposed as a noise-robust model for relation extraction and has shown promising results. CETA contains two discrepant classifiers that share a single encoder. The focus of CETA is to train the classifiers only in instances where both classifiers have reached a consensus. To achieve consensus, CETA augments the standard cross entropy loss to include predictions from both classifiers and uses the Wasserstein distance \cite{kantorovich2006translocation} as a secondary criterion. 
\subsection{Deep Ensembles}
Deep ensembles have been shown to generally exhibit robustness as compared to singular models and reduce overfitting \cite{ensemble2022105151}. To that end, we hypothesize that ensembles may excel in noisy classification tasks due to the presence of label noise in the training data, which can cause individual models to learn false correlations between features and labels. By training multiple classifiers and combining their predictions, each model can develop a unique representation of the input data, leading to a more robust classification boundary. While ensembles have been previously proposed for data and label noise detection \cite{wheway2001boosting, sluban2014ensemble,chong-etal-2022-detecting}, their performance as a method of robust text classification with noisy labels has not been established. 

We formally define ensembles as follows: Given $m$ classifiers $C_1, C_2, ..., C_m$, each classifier produces probabilities $P_{c_i}$ on a clean test set $T$, an ensemble of the predictors averages the probabilities of each predictor such that $P_{ensemble} = \sum_{i=1}^{m} \frac{P_{C_i}}{m}$. It should be noted that each ensemble member is trained on either the same noisy training set or a randomly selected subset of the noisy training set, depending on the employed technique, which are are described below. Nevertheless, in all scenarios, each member is evaluated on the same clean test set. 
We experiment with three types of ensembles:
\begin{enumerate}
    \item \textbf{Homogeneous Ensembles} Ensembles that aggregate predictions from the same type of classifier (i.e. vanilla BERT with early stopping), trained with different initializations and hyperparameters.  
    \item \textbf{Heterogeneous Ensembles} Ensembles that aggregate predictions from different types of classifiers. In our experiments, we use vanilla BERT, co-teaching, and CETA as the heterogeneous classifiers in the ensemble. 
    \item \textbf{Boosting} Ensembles that aggregate predictions from the same type of classifier (i.e. vanilla BERT with early stopping), but each classifier is trained on a different subset of the training data. 
\end{enumerate}

\subsection{Noise Cleaning Based on Fine-Tuned Model Loss} \label{sec:noise_cleansing}
We use the pre-trained language model's ability to identify noisy labels as a way to clean the training set by removing instances with potential label noise. This involves fine-tuning BERT on noisy task-specific training data and evaluating model loss on each instance. Training instances that have a loss higher than the selected threshold are excluded from the training set used to train the final classifier. We tune the loss threshold on a noisy validation set.  

To avoid biasing or overfitting the model when computing loss on the same set used for fine-tuning, we employ an N-fold process to calculate the loss only on held-out data points\footnote{A similar approach is briefly described in \cite{pervasive_errors} for estimating noise characterization in the confident learning framework.}. The process is outlined in Algorithm \ref{cleansing_algorithm}. 
In summary, we fine-tune the model using a subset of the noisy training set and use the model to identify and remove noisy samples from the held-out validation set using a fixed loss threshold\footnote{The loss threshold is a hyperparameter that we tune beforehand.}. The process is repeated separately N times using disjoint validation sets to clean the complete training set.

\begin{algorithm}
\caption{Noise Cleaning Algorithm}\label{cleansing_algorithm}
\begin{algorithmic}[1]
\State \textbf{Input:} Noisy train set $T$, loss threshold $t$, number of folds $f$
\State \textbf{Output:} Cleaned train set $T_{\text{clean}}$
\State Divide $T$ into $f$ validation subsets: $V_{1}, \ldots, V_{f}$
\For{$i = 1$ to $f$}
\State $T_{i} = T \setminus V_{i}$
\State Train a fine-tuned model $M_{i}$ on $T_{i}$
\State Evaluate the model loss $L_{V_{i}}$ on $V_{i}$
\State $T_{\text{clean},i} = V_{i}[L_{V_{i}} < t]$
\EndFor
\State $T_{\text{clean}}$ = $\bigcup_{i=1}^{f} T_{clean,i}$
\State \textbf{return} $T_{\text{clean}}$
\end{algorithmic}
\end{algorithm}

\section{Experiments and Results}
All of the methods evaluated in these experiments incorporate early stopping on noisy validation set as done by \citet{bert2022robust}. We use a noisy validation set because obtaining a clean validation set is often not feasible in practice. Moreover, \citet{bert2022robust} show that using a noisy validation set for early stopping is more or less as effective as using a clean validation set. 

\subsection{Hyperparameters}

The number of training steps is optimally set to 3000\footnote{If the validation accuracy does not improve beyond a certain patience level, we employ early stopping to prematurely halt the training process for all experiments.} unless we are required to vary hyperparameter settings for homogeneous ensembles.
For homogeneous ensembles, we cycle through a combination of the following hyperparameters: the number of training steps = [2000, 3000, 4000, 5000, 6000], learning rate = [0.0002, 0.0004, 0.0005, 0.00001, 0.00002, 0.00003, 0.00004, 0.00005], patience (for early stopping) = [25, 30, 40, 50], warm-up steps = [0, 1, 5, 7, 10], weight decay = [0.1, 0.001, 0.0001], and drop rate = [0.1, 0.25, 0.5, 0.8]. 

For other experiments that do not explicitly require us to vary hyperparameters, we fix the following hyperparameters for the African language datasets, training steps = 3000, learning rate = 0.00005, patience = 25, drop rate = 0.1, warm-up steps = 0, weight decay = 0.1. We fix the following hyperparameters for the English language datasets, training steps = 3000, learning rate = 0.00002, patience = 25, drop rate = 0.25, warm-up steps = 0, weight decay = 0.1. For boosting related experiments, we experiment with two training data subset sizes: 50\% of the total training data and 80\% of the total training data. For heterogeneous ensembles, we aggregate predictions from the following three classifiers: vanilla BERT, co-teaching, and CETA.

\subsection{BERT Models}
We use \textit{bert-base-uncased}\footnote{https://huggingface.co/bert-base-uncased} as the backbone for our English language datasets: TweetNLP and SNLI. We use \textit{bert-base-multilingual-cased}\footnote{https://huggingface.co/bert-base-multilingual-cased} for our African language datasets: Yor\`ub\'a and Hausa.

\begin{table}[t]
    \centering
    \scalebox{0.9}{
    \begin{tabular}{|c|c|c|}
        \hline
        & Hausa & Yor\`ub\'a\\
        \hline
        \rowcolor{gray!50}
        \multicolumn{1}{|c}{} & \multicolumn{2}{c|} {Clean Data} \\
        \hline
        Vanilla BERT & $82.67 \pm 0.73$ & $76.23 \pm 0.28$\\
        \hline
        \rowcolor{gray!50}
        \multicolumn{1}{|c}{} & \multicolumn{2}{c|} {Noisy Data} \\
        \hline
        Vanilla BERT & $46.98 \pm 1.01$ & $64.72 \pm 1.45$\\
        Co-Teaching & $\textbf{48.11} \pm 1.71$ & $64.38 \pm 0.98$\\
        \hline
        CETA & $^*\textbf{49.31} \pm 0.31$ & $^*\textbf{68.07} \pm 0.18$\\
        \hline
        HME  & $46.39 \pm 0.21$& $\textbf{67.28} \pm 0.81$\\
        HTE & $ \textbf{48.28} \pm 0.19$& $ \textbf{67.81} \pm 0.73$\\
        Boosting & $\textbf{47.13} \pm 0.42$ & $\textbf{67.63} \pm 1.26$ \\
        \hline
        NC & $\textbf{47.18} \pm 0.22 $ & $62.17 \pm 0.54$ \\
        \hline
    \end{tabular}
    }
    \caption{A comparison of proposed methods against baselines on datasets with feature-dependent label noise. \textbf{HME}: Homogeneous ensembles \textbf{HTE}: Heterogeneous ensembles. Boosting: Ensembles of different random subsets from the train set. \textbf{NC}: Noise Cleaning. Average accuracy is reported with a standard deviation from 5 runs of each experiment.}
    \label{tab:summ1}
\end{table}

\subsection{Loss threshold}
To select a loss threshold for noise-cleaning as described in section \ref{sec:noise_cleansing}, we experiment with different cut-off points in the following interval $[6.0, 8.0]$. We use only a noisy validation set to select the loss threshold. Data points whose loss exceeds the fixed loss threshold are excluded from the training set, effectively `cleaning' the noisy training set to a certain extent. Note that we only report results on the loss threshold that produces the most optimal accuracy on the noisy validation set. The cleaned training set is once again used to train a vanilla BERT model, at which point we can evaluate how well the noising scheme performed.

\begin{figure}
  \centering
  \begin{subfigure}{0.23\textwidth}
    \centering
    \includegraphics[width=\textwidth]{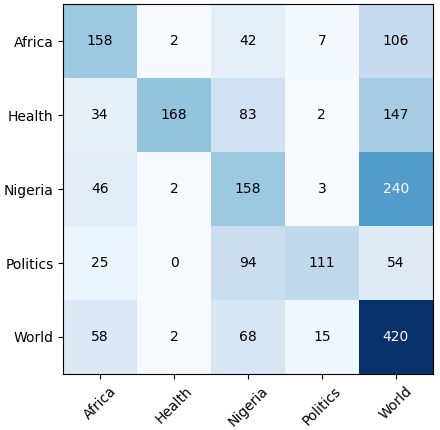}
    \caption{Hausa: before}
    \label{hausa_before}
  \end{subfigure}\hfill
  \begin{subfigure}{0.23\textwidth}
    \centering
    \includegraphics[width=\textwidth]{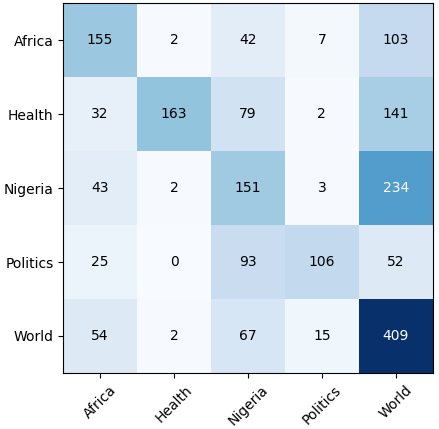}
    \caption{Hausa: after}
    \label{hausa_after}
  \end{subfigure}\\
  \begin{subfigure}{0.24\textwidth}
    \centering
    \includegraphics[width=\textwidth]{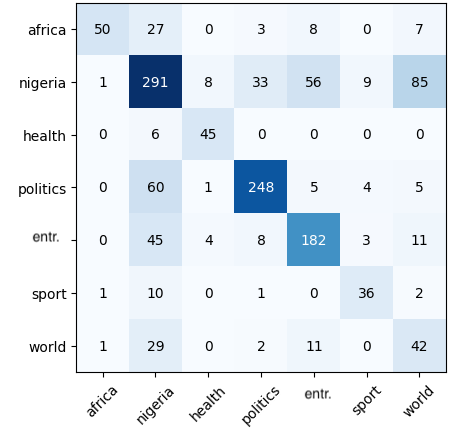}
    \caption{ Yor\`ub\'a: before}
    \label{yoruba_before}
  \end{subfigure}\hfill
  \begin{subfigure}{0.24\textwidth}
    \centering
    \includegraphics[width=\textwidth]{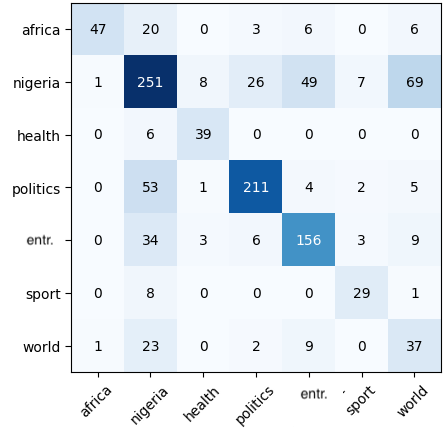}
    \caption{ Yor\`ub\'a: after}
    \label{yoruba_after}
  \end{subfigure}
  \caption{Noise matrices for Hausa and Yor\`ub\'a showing noise distribution before and after noise cleaning.}
  \label{fig:hausa_and_yoruba_b_and_after}
\end{figure}

\subsection{Results}

\begin{table*}[ht]
    \centering
    \scalebox{0.9}{
    \begin{tabular}{|c|c|c|c|c|c|c|}
        \hline
        & \multicolumn{3}{c}{TweetNLP} & \multicolumn{3}{|c|}{SNLI}\\
        \hline
       Noise Level & 10\% & 20\% & 30\% & 10\% & 20\% & 30\% \\
        \hline
        \rowcolor{gray!50}
        \multicolumn{1}{|c}{} & \multicolumn{6}{c|} {Clean Data} \\
        \hline
        Vanilla BERT & \multicolumn{3}{c|}{$91.03 \pm 0.81$} & \multicolumn{3}{c|}{$85.03 \pm 0.16$}\\
        \hline
        \rowcolor{gray!50}
        \multicolumn{1}{|c}{} & \multicolumn{6}{c|} {Noisy Data} \\
        \hline
        Vanilla BERT & $82.08 \pm 0.03$ & $74.45 \pm 0.65$ & $72.96 \pm 1.42$  & $84.79 \pm 0.87$ & $\textbf{83.83} \pm 1.01$ & $82.01 \pm 0.21$\\
        Co-Teaching & $81.31 \pm 0.11$ & $73.68 \pm 0.04$ & $72.41 \pm 0.71$ & $84.27 \pm 0.15$ & $83.10 \pm 1.20$ & $80.99 \pm 0.04$\\
        \hline
        CETA & $81.00 \pm 1.81$ & $72.40 \pm 1.01$ & $72.13 \pm 0.71$ & $84.24 \pm 0.01$ & $82.67 \pm 0.21$ & $81.02 \pm 0.27$\\
        \hline
        HME & $81.81 \pm 0.05$ & $74.08 \pm 0.03$ & $72.53 \pm 0.01$ & $85.02 \pm 0.12$ & $83.76 \pm 0.10$ & $81.99 \pm 0.26$\\
        HTE & $79.13 \pm 0.32$ & $\textbf{74.90} \pm 0.51$ & $72.32 \pm 0.97$ & $84.75 \pm 0.34$ & $83.64 \pm 1.11$ & $81.16 \pm 0.97$\\
        Boosting & $\textbf{82.53} \pm 0.01$ & $74.27 \pm 0.15$ & $\textbf{73.52} \pm 3.32$ & $\textbf{85.38} \pm 0.45$  & $83.80 \pm 0.81$ & $\textbf{82.06} \pm 0.41$ \\
        \hline
        NC & $80.94 \pm 0.09$ & $74.55 \pm 0.45$ & $72.65 \pm 0.19$ & $85.13 \pm 0.05$ & $\textbf{84.00} \pm 0.01$ & $\textbf{82.97} \pm 1.09$ \\
        \hline
    \end{tabular}
    }
    \caption{A comparison of proposed methods against baselines on TweetNLP and SNLI datasets noised to various levels. HME: Homogeneous ensembles HTE: Heterogeneous ensembles. Boosting: Ensembles of different random subsets from the training set. Average accuracy is reported with the standard deviation from 5 runs of each experiment. }
    \label{tab:summ2}
\end{table*}

\subsection{Feature-dependent label noise}
Table \ref{tab:summ1} shows the results of baseline models and the proposed approaches on datasets containing feature-dependent label noise: Hausa and Yor\`ub\'a. 

First, we observe that co-teaching and noise cleaning do not consistently improve performance compared to vanilla BERT. CETA, on the other hand, improves performance by around 3 absolute percentage points on both datasets. The homogeneous ensemble method does not consistently improve either, but we do observe consistent gains using heterogeneous ensembles and boosting.

Figure \ref{fig:hausa_and_yoruba_b_and_after} show the noise distribution in the training set before and after applying the noise cleaning procedure in both datasets. Note that the noise-cleaning method results in the removal of both noisy and clean instances, which leads to the total noise level not being considerably reduced. Overall, we we do not observe a larger reduction in noise level in either dataset. After noise cleaning, we have 31\% label noise in Yor\`ub\'a compared to 33\% before noise cleaning, with only a 2\% reduction in noise. For Hausa, the noise level after cleaning is similarly reduced by 3\% (47\% compared to 50\% before cleaning). In summary, we do not find the noise-cleaning method to be an efficient error detector for feature-dependent label noise, as compared to the other noise-robust we use. This is inconsistent with the result in \citet{chong-etal-2022-detecting}, where they show that language models are suitable for label error detection. However, they also report that an  \textit{\textbf{ensemble of large}} pre-trained language models is a better error detector than a smaller individual pre-trained model, and in both cases, while models may have good error detection performance, the performance in the underlying task is not necessarily improved. 

\subsection{Pseudo-real-world label noise}

\begin{table*}[h]
  \centering
  \scalebox{0.8}{
  \begin{tabular}{ l p{4.5in} l l }
    \hline
    Dataset & Text & Noisy Label & Actual Label \\ 
    \hline
    SNLI & (1) Young man wearing a blue jacket, green shirt and denim jeans is photographed by person in beige jacket and burgundy pants while four onlookers watch on an expanse of sand.\textcolor{red}{<!SEP!>} The people are ignoring the man getting photographed. & \textcolor{blue}{No Relationship} & \textcolor{red}{\sout{Contradiction}} \\
    SNLI & (2) A man wearing a black t-shirt is playing seven string bass a stage.\textcolor{red}{<!SEP!>} The man is playing an old guitar. & \textcolor{blue}{Contradiction} & \textcolor{red}{\sout{No Relationship}} \\
    SNLI & (3) Many children are sitting in a classroom watching a woman in the front.<!SEP!>The woman is teaching the children & \textcolor{blue}{Entailment} & \textcolor{red}{\sout{No Relationship}} \\
    TweetNLP & (1) Reading harry potter in bed! waiting for the new \textcolor{red}{south} park to come on & \textcolor{blue}{ADJ} & \textcolor{red}{\sout{NOUN}}\\
    TweetNLP & (2) @USER: I'm not insulted, \textcolor{red}{at} all, trust me. I'm seeking to understand you and your video. :) & \textcolor{blue}{DET} & \textcolor{red}{\sout{ADP}}\\
    TweetNLP & (3) Chicagoan early voters in Uptown even get brownies and entertainment while waiting for a dozen people to do number \textcolor{red}{page} ballots. & \textcolor{blue}{ADJ} & \textcolor{red}{\sout{NOUN}}\\ 
    \hline
  \end{tabular}
  }
  \caption{Samples from SNLI and TweetNLP with pseudo-real-world noise injection, highlighting the complexity and potential ambiguity of these tasks.}
  \label{table:noise}
\end{table*}
Table \ref{tab:summ2} shows the results on datasets containing pseudo-real-world label noise, TweetNLP, and SNLI, with three levels: $10\%$, $20\%$, and $30\%$. In these datasets, we observe that performance drops significantly with increased noise levels in TweetNLP, but only small drops in performance are observed in SNLI. We hypothesize that this potentially reflects the inherent difficulty in the natural language inference task, and the gold labels may already be ambiguous even before applying the noising scheme. Table \ref{table:noise} shows samples from both SNLI and TweetNLP datasets before and after injecting noisy labels. In many cases, particularly in SNLI, the given example is rather ambiguous and both labels can be suitable. These are also cases where there are high inter-annotator disagreements.  

In terms of noise handling techniques, we observe that all approaches generally do not produce large gains in performance compared to vanilla BERT. Furthermore, many approaches result in slightly worse performance compared to the baseline. Boosting seems like the most robust technique, as it maintains baseline performance at least, while also being effective against feature-dependent label noise. Noise cleaning in this category obtained mixed results. Surprisingly, CETA does not excel over other methods in this particular category. Although it was specifically designed to address feature-dependent label noise, its performance is somewhat inferior to the vanilla BERT baseline when dealing with realistic label noise. We surmise that this type of artificial noise is more challenging as it's based on actual human errors, and may even reflect intrinsic ambiguities in the task, which makes it harder to detect through automatic approaches.

\section{Conclusions}
In this paper, we described experiments for evaluating different label noise handling techniques within the framework of BERT text classification. We evaluated some multi-network training approaches (i.e. co-teaching and CETA), different types of ensembles (homogeneous, heterogeneous, and boosting), and a noise cleaning technique and compared them with a vanilla BERT fine-tuned model with early stopping. We used two datasets that contain feature-dependent label noise from automatic labeling, as well as two datasets with synthetic pseudo-real-world label noise obtained by considering multiple annotations.  

 For feature-dependent label noise, the recently proposed Consensus Enhanced Training Approach (CETA) shows the most promising results compared to the baselines. Some ensembling techniques, such as boosting, can also improve performance compared to the baselines but do not provide the level of robustness achieved via CETA. 

 While pre-trained language models have been shown previously to have the potential to detect label errors through out-of-sample loss, our results indicate that using this technique to automatically clean the data does not result in improved performance compared to using the noisy set. This may suggest that removing label errors is not necessarily a good approach for handling label noise; rather, error detection can be used to identify noisy labels for manual correction. 

The synthetic pseudo-real-world category of label noise appears to be more challenging as the noise represents actual human errors, which could be an indication of inherent ambiguities in the task itself. Our experiments show that most techniques do not improve performance compared to the baselines. Furthermore, for a dataset like SNLI, which is known to be challenging even for human annotators, the presence of label noise does not reduce the performance to a great extent compared to the other datasets. This may suggest that the noising scheme is compatible with the inherent difficulty or label ambiguity of the task, and any attempts to detect or discard the noise will not necessarily improve the performance using stringent metrics such as accuracy. Recent efforts to embrace annotator disagreements and incorporate them in the training process \cite{zhang2021learning} rather than relying on a single gold label may be more suitable to handle this kind of labeling inconsistencies. 

Overall, the results indicate that handling realistic label noise in text classification remains a challenging task, and none of the noise-handling techniques examined so far has shown consistent performance improvements across multiple datasets.

\section*{Limitations}

The work described in this paper is limited by the small number of datasets that contain both noisy and clean versions in the text classification domain, which are needed for evaluating noise-handling methods. While we observed positive results from at least two approaches, any conclusions we make about their effectiveness are drawn from a sample of two datasets, and may not necessarily generalize to other cases. For the pseudo-real-world label noise category, it is unclear whether the noise represents true errors or inherent ambiguity in the task. The mixed results we observe could also be a result of ambiguities in the presumed `clean' test set. 

\bibliography{anthology,custom}
\bibliographystyle{acl_natbib}
\end{document}